\documentclass[preprint]{elsarticle}
\usepackage{amsmath,amsfonts}
\usepackage{epsfig}
\usepackage{graphicx}
\usepackage{color}
\usepackage{soul}
\usepackage{booktabs,caption}
\usepackage[flushleft]{threeparttable}
\usepackage{booktabs}
\usepackage{multirow}
\usepackage[plain,noend]{algorithm2e}
\usepackage{subfigure}
\usepackage{xcolor}
\usepackage{dcolumn}
\usepackage{comment}
\usepackage{tabu}
\usepackage{hyperref}

\journal{Computers \& Operations Research}

\begin{document}

\begin{frontmatter}

\title{Cost-sensitive feature selection for Support Vector Machines}

\author[uss,us]{S. Ben\'itez-Pe\~na\corref{mycorrespondingauthor}}
\cortext[mycorrespondingauthor]{Corresponding author}
\ead{sbenitez1@us.es}

\author[uss,us]{R. Blanquero}
\ead{rblanquero@us.es}

\author[uss,us]{E. Carrizosa}
\ead{ecarrizosa@us.es}

\author[uss,uca]{P. Ramirez-Cobo}
\ead{pepa.ramirez@uca.es}

\address[uss]{IMUS. Universidad de Sevilla. 41012 Sevilla. Spain}
\address[us]{Departamento de Estad\'istica e Investigaci\'on Operativa. Universidad de Sevilla. 41012 Sevilla. Spain}
\address[uca]{Departamento de Estad\'istica e Investigaci\'on Operativa. Universidad de C\'adiz. 11510 Puerto Real, C\'adiz. Spain}

\begin{abstract}
Feature Selection is a crucial procedure in Data Science tasks such as Classification, since it identifies the relevant variables, making thus the classification procedures more interpretable, cheaper in terms of measurement and more effective by reducing noise and data overfit.
The relevance of features in a classification procedure is  linked to the fact that misclassifications costs are frequently asymmetric, since false positive and false negative cases may have very different consequences. However,
off-the-shelf Feature Selection procedures seldom take into account such cost-sensitivity of errors.

In this paper we propose a mathematical-optimization-based Feature Selection procedure embedded in one of the most popular classification procedures, namely,  Support Vector Machines,  accommodating asymmetric misclassification costs.
The key idea is to replace the traditional margin maximization by minimizing the number of features selected, but imposing upper bounds on the false positive and negative rates. {The problem is written as an integer linear problem plus a quadratic convex problem for Support Vector Machines with both linear and radial kernels.}

The reported numerical experience demonstrates the usefulness of the proposed Feature Selection procedure. Indeed, our results on benchmark data sets show that a substantial decrease of the number of features is obtained, whilst the desired trade-off between false positive and false negative rates is achieved.
\end{abstract}


\end{frontmatter}

\section{Introduction}
\label{introduction}
Supervised Classification is one of the most important tasks in Data Science, e.g. \cite{bertsimas2016tae,provost2013data}, full of challenges from a Mathematical Optimization perspective, e.g. \cite{bartlett2006convexity,
ben2011chance,
boct2011optimization,
bradley1999mathematical,
carrizosa2016strongly,
carrizosa2017clustering, carrizosa2013supervised,
corne2012synergies,marron2007distance,
meisel2010synergies,panagopoulos2016constrained,plastria2012minmax,richtarik2016parallel,kxw018,shen2003psi}.
 In its most basic version,  we are given a set $I$ of individuals, each  represented by a vector $(x_i, y_i)$, where $x_i \in \mathbb{R}^N$  is the so-called feature vector, and $y_i \in \mathcal{C}=\left\{ -1,1\right\}$  is the membership of individual $i$. A classifier $\Psi,$  i.e., a function $\Psi: \, \mathbb{R}^N \, \longrightarrow \, \mathcal{C}$, is sought to assign labels $c \in \mathcal{C}$  to incoming individuals for which the feature vector $x$ is known but the label $y$ is unknown and estimated through $\Psi(x).$

The different classification procedures differ in the way the classifier $\Psi$ is obtained from the data set $I$. A frequent approach consists of reducing the search of the classifier to the resolution of an optimization problem, see \cite{carrizosa2013supervised}. This is the case, among many others, of the state-of-the-art classifier for binary classification, known as Support Vector Machines (SVM), \cite{carrizosa2013supervised,cristianini2000introduction, vapkinnature, vapnik1998statistical}, addressed in this paper.

In SVM \emph{with linear kernel}, $\Psi$ takes the form
\begin{equation}
\label{eq:score1}
\Psi(x) = \left\{
\begin{array}{rl}
1, & \mbox{if $\boldsymbol{w}^\top x + \beta \geq 0$ } \\
-1, & \mbox{else},
\end{array}
\right.
\end{equation}
\noindent where $\boldsymbol{w}\in\mathbb{R}^N$ and $\beta\in\mathbb{R}$ are  obtained as the optimal solution of the following  convex quadratic programming formulation with linear constraints
\begin{equation}
\label{eq:svm1}
\begin{array}{lll}
  \min_{\boldsymbol{w}, \beta, \xi} & \boldsymbol{w}^\top \boldsymbol{w} + C\sum_{i \in I} \xi_i& \\
  s.t. & y_i(\boldsymbol{w}^\top x_i + \beta) \geq 1 - \xi_i,& i \in I \\
   & \xi_i \geq 0& i \in I.
\end{array}
\end{equation}
Here $C>0$ is the \textit{regularization parameter}, which needs to be tuned, and $\xi_i \geq 0$ is a penalty associated to misclassifying individual $i$ in the so-called training sample $I$.

An apparently innocent extension of (\ref{eq:score1}) is given by
 \begin{equation}
\label{eq:score2}
\Psi(x) = \left\{
\begin{array}{ll}
1, & \mbox{if $\boldsymbol{w}^\top \phi(x) + \beta \geq 0$ } \\
{-1}, & \mbox{else},
\end{array}
\right.
\end{equation}
\noindent where $\phi: \mathbb{R}^N \rightarrow \mathcal{H}$ maps the original $N$ features into a vector space of higher dimension, and $\boldsymbol{w}$ and $\beta$ are obtained by solving an optimization problem  formally identical to (\ref{eq:svm1}), but taking place in the space $\mathcal{H}$ instead of $\mathbb{R}^N$,
\begin{equation}
\label{eq:svm2}
\begin{array}{lll}
  \min_{\boldsymbol{w}, \beta, \xi} & \boldsymbol{w}^\top \boldsymbol{w}+ C\sum_{i \in I} \xi_i& \\
  s.t. & y_i(\boldsymbol{w}^\top \phi(x_i) + \beta) \geq 1 - \xi_i,& i \in I \\
   & \xi_i \geq 0& i \in I.
\end{array}
\end{equation}
In this case, the classifier is usually obtained by solving, instead of (\ref{eq:svm2}), its dual,
\begin{equation}
\label{eq:svm3}
\begin{array}{lll}
  \max_{\alpha} & \sum_{i \in I} \alpha_i  - \frac{1}{2}\sum_{i,j \in I} \alpha_i y_i \alpha_j y_j K(x_i,x_j)  \\
  s.t. &  \sum_{i \in I} \alpha_i y_i = 0\\
   & 0 \leq \alpha_i \leq \frac{C}{2}, &  i \in I,
\end{array}
\end{equation}
\noindent where $K(x,x')=\phi(x)^\top\phi(x')$ is the so-called \textit{kernel function}.
From the optimal solution to (\ref{eq:svm3}) and taking into account the complementarity slackness conditions, $\boldsymbol{w}$ and $\beta$ in  (\ref{eq:score2}) are obtained.
In particular,
\begin{eqnarray}
\label{eq:recovery}
\boldsymbol{w}^\top \boldsymbol{w} & = & \sum_{i,j \in I} \alpha_i y_i \alpha_j y_j K(x_i,x_j), \\
\label{eq:recoverscore}
\boldsymbol{w}^\top \phi(x)& = & \sum_{i \in I} \alpha_i y_i K(x_i,x).
\end{eqnarray}
See e.g. \cite{carrizosa2013supervised,cristianini2000introduction, vapkinnature, vapnik1998statistical} for details.

The classifier uses all the features involved in the problem, both in (\ref{eq:score1}) and (\ref{eq:score2}), which may be rather problematic if measuring the features involve some non-negligible costs. This is particularly relevant when the dimension $N$ of the data set is large. It is then advisable to perform Feature Selection (FS), \cite{aytug2015feature,bertolazzi2016integer,bradley1998feature,carrizosa2011detecting,fung2004feature,guyon2003introduction,le2015feature,maldonado2009wrapper,maldonado2011simultaneous,weston2000feature}, in order to reduce the set of features and obtain an appropriate trade-off between classification accuracy and sparsity.

An amount of different FS procedures are found in the literature, some independent of the classification procedure (FS is performed in advance, based e.g. on the correlation between each feature and the label) and others embedded in the classification procedure, {like the Holdout SVM (HOSVM), \cite{MALDONADO20092208}, Kernel-Penalized SVM (KP-SVM), \cite{MALDONADO2011115}, or the methods presented in \cite{Chan2007DCR12734961273515} or \cite{GHADDAR2018993}}. {Also, one can minimize the number of relevant features or even their cost, as in \cite{MALDONADO2017656}.  The embedded method together with whichever of the previous optimization schemes is the approach considered in this paper}, since we aim to obtain a SVM-based classifier, and, at the same time, perform the selection of the features.
The core idea is the optimization problem to be solved:  instead of maximizing the margin, as in the traditional SVM, we seek the classifier with lowest number of features {(or cost of the features)}, {but without damaging too much the original performance.} {In order to be able to control the classifier's performance, we will make use of constraints as in \cite{CSVM}. Specifically, the formulation of the constrained SVM with linear kernel is
\begin{equation}
\label{eq:csvm}
\begin{array}{lll}
  \min_{\boldsymbol{w}, \beta, \xi} & \boldsymbol{w}^\top \boldsymbol{w}+ C\sum_{i \in I} \xi_i& \\
  s.t. & y_i(\boldsymbol{w}^\top x_i + \beta) \geq 1 - \xi_i,& i \in I \\
   & 0 \leq \xi_i \leq M_1(1-\zeta_i)& i \in I\\
   & \mu(\zeta)_\ell \geq \lambda_\ell & \ell \in L\\
   & \zeta_i \in \{0,1\} & i \in I,
\end{array}
\end{equation}
{where $M_1$ is a large number.}

{In essence, this is simply the formulation for the SVM with linear kernel, to which performance constraints have been added: $\mu(\zeta)_\ell \geq \lambda_\ell${, where $\mu(\zeta)_\ell$ are different performance measures, forced to take values above thresholds $\lambda_\ell$, and $\zeta_i$ are new binary variables that check whether sample $i$ is counted as correctly classified}. See \cite{CSVM} for the details. Its}
(partial) dual formulation is
\begin{equation}
\label{eq:csvm2}
\begin{array}{lll}
  \min_{\alpha, \beta, \xi, \zeta} & \sum\limits_{i,j \in I} \alpha_iy_i\alpha_jy_jK(x_i,x_j)+ C\sum_{i \in I} \xi_i& \\
  s.t. & y_i(\sum\limits_{j \in I}\alpha_j y_j K(x_j,x_i) + \beta) \geq 1 - \xi_i,& i \in I \\
  & \sum\limits_{i \in I}\alpha_i y_i = 0\\
  & 0 \leq \alpha_i \leq C/2 & i \in I\\
   & 0 \leq \xi_i \leq M_1(1-\zeta_i)& i \in I\\
   & \mu(\zeta)_\ell \geq \lambda_\ell & \ell \in L\\
   & \zeta_i \in \{0,1\} & i \in I.
\end{array}
\end{equation}
{As before, this is similar to the standard partial dual formulation of the SVM with general kernel and constraints in the performance measures, as in (\ref{eq:csvm}).} For more information about how formulation~(\ref{eq:csvm2}) is obtained, the reader is referred to the Appendix.} Note that, while mathematical optimization problems addressed in the statistical literature are, traditionally, as (\ref{eq:svm1}) or (\ref{eq:svm3}), nonlinear programs in continuous variables, our approach involves integer variables, which define harder optimization problems. However, Integer Programming has shown to be rather competitive thanks to the impressive advances in (nonlinear) integer programming solvers, as demonstrated in recent papers addressing different topics in data analysis, \cite{bertsimas2016best,bertsimas2014least,carrizosa2011detecting,carrizosa2016strongly,carrizosa2017clustering,carrizosa2016sparsity}.

The remainder of the paper is structured as follows. In Section~\ref{FS} we present the new FS methodology for SVM. {For either linear or nonlinear kernels, we reduce the optimization problem to solving a standard linear integer program plus, eventually, a quadratic convex problem.}
The performance of our FS approach is empirically tested under different experiments described in Section~\ref{Experimentdescription}. The results of those experiments are shown in Section~\ref{Results}. {Comparisons between the use of linear and radial kernels, and between the standard linear SVM with and without embedded FS are also provided.} The paper ends with conclusions and possible extensions in Section~\ref{se:concluding}.

\section{Cost-sensitive Feature Selection}
\label{FS}

In this section we present a novel linear formulation for SVM where classification costs are modeled via certain constraints, and where, in addition, a FS approach is embedded in such a way that only the relevant features are considered.

In order to cope with classification costs, first  we recall some performance measures, namely,
\begin{itemize}
\item TPR (True Positive Rate): $P(\boldsymbol{w}^\top {X} + \beta > 0 | {Y} = +1)$
\item TNR (True Negative Rate): $P(\boldsymbol{w}^\top {X} + \beta < 0 | {Y} = -1)$
\item Acc (Accuracy):  $P({Y}(\boldsymbol{w}^\top {X} + \beta) > 0),$
\end{itemize}
{where $P(\boldsymbol{w}^\top {X} + \beta > 0 | {Y} = +1)$ and $P(\boldsymbol{w}^\top {X} + \beta < 0 | {Y} = -1)$ denote, respectively, the probability of classify correctly a positive or negative labeled instance, and $P({Y}(\boldsymbol{w}^\top {X} + \beta) > 0)$ is probability of classify correctly a given instance.}

The objective is to perform classification using a reduced set of features, in such a way that certain constraints over the performance, such as $TPR \geq \lambda_1$ or $TNR \geq \lambda_{-1}$ (for threshold values $\lambda_1$, $\lambda_{-1} \in [0,1]$), are fulfilled.

Note that the pair {$(X,Y)$} is a random vector with unknown distribution from which a sample {$\left\{(x_i,y_i)\right\}_{i \in I}$} is generated. This implies that $TPR$ and $TNR$ should be estimated from sample data. This leads to the empirical constraints $\widehat{TPR} \geq \lambda_1^*$ and $\widehat{TNR} \geq \lambda_{-1}^*$, for $\lambda_1^*\geq \lambda_1$ and $\lambda_{-1}^* \geq \lambda_{-1}$, where the performance measures are replaced by their sample estimates.
{Two possible choices, which shall be explored in this work, are
\begin{equation}
\label{eq:HoeffornotNot}
\begin{array}{lll}
   \lambda_1^* & = & \lambda_1 \\
   & \text{and} & \\
  \lambda_{-1}^* & = & \lambda_{-1}, \\
\end{array}
\end{equation}
or the more conservative approach based on Hoeffding inequality,
 \begin{equation}
\label{eq:HoeffornotHoeff}
\begin{array}{lll}
   \lambda_1^* & = & \lambda_1 + \sqrt{\dfrac{-\log \alpha}{2|I_+|}} \\
   & \text{and} & \\
  \lambda_{-1}^* & = & \lambda_{-1} + \sqrt{\dfrac{-\log \alpha}{2|I_-|}},  \\
\end{array}
\end{equation}
where $\alpha$ is the significance level for the hypothesis test whose null hypothesis is either $TPR \leq \lambda_1$ or $TNR \leq \lambda_{-1}$.} See \cite{CSVM} for more details.

Note that it is straightforward to extend our results to the case in which measurement costs are associated with the features, as in e.g.  \cite{carrizosa2008multi}, and then the minimum-cost feature set is sought instead.

\subsection{The cost-sensitive FS procedure}\label{FSProc}
Assume that we have a linear kernel, i.e., the kernel $K$ in (\ref{eq:svm3}) is given by $K(x,x') = x^\top x',$ and thus the SVM with all features is obtained by solving (\ref{eq:svm1}).
We state the feature selection problem as a Mixed Integer Linear Program.
Consider an auxiliary variable $\zeta_i$ that {in case of being equal to 1, the instance $i$ is counted as correctly classified}.
Hence, estimates of  TPR and TNR from  sample ${I}$ {have lower bounds} $\widehat{TPR} {\geq} \sum_{i \in I} \zeta_i (1+y_i )/{\sum_{i \in I} (1+y_i)}$ and $\widehat{TNR} {\geq} \sum_{i \in I} \zeta_i (1-y_i )/{\sum_{i \in I} (1-y_i)}$, respectively.
Associated with each feature $k$, $1\leq k \leq N$, we define the variable $z_k$ taking the value 1 if feature $k$ is selected for classifying, and 0 otherwise.
 Hence, the optimization problem that defines a {linear classifier (hyperplane)} taking into account the classification rates and in which a {cost-based} FS procedure is integrated is given by

 $$
\left.
\begin{array}{lll}
\min_{\boldsymbol{w},\beta,z,\zeta} & \sum\limits_{k=1}^{N} {c_k} z_k & \\
s.t. & y_i(\boldsymbol{w}^\top x_i + \beta) \geq 1 - M_2(1-\zeta_i),& \forall i \in I \\
     &  \sum_{i \in I} \zeta_i (1-y_i )  \geq \lambda_{-1}^* {\sum_{i \in I} (1-y_i)}&\\
     & \sum_{i  \in I} \zeta_i (1+y_i) \geq \lambda_{1}^*\sum_{i \in I}(1+y_i )&\\
     & |{w}_k| \leq M_3z_k & \forall k \in {1,\ldots,N}\\
     & \zeta_i \in \{0,1\}& \forall i \in I\\
     & z_k \in \{0,1\}& \forall k \in {1,\ldots,N}
\end{array}
\right.
\qquad (P1)
$$

\noindent
where $M_2$ and $M_3$ are sufficiently large numbers. {Also, $c_k$ is the cost associated to the $k$-th feature, so we perform the FS by reducing the overall cost of the features. The case $c_k = 1 \quad \forall k$, is the standard FS in which the number of features selected is minimized.}

Let us discuss the rationality of the formulation $(P1)$. The overall cost associated with the features used for classifying is to be minimized in the objective. The first constraint identifies which individuals are counted as correctly classified, since, as soon as $\zeta_i = 1,$ the score $\Psi(x_i)$ is forced to be $\Psi(x_i) \geq 1$ (if $y_i=1$) or $\Psi(x_i)\leq -1$ (if $y_i=-1$). Furthermore,  the constant $\sum_{i \in I} (1-y_i)$ is equal to two times the cardinality of the set $\{i \in I: y_i=-1\}$, whereas  $\sum_{i \in I} \zeta_i(1-y_i)$ yields two times the number of individuals counted as correctly classified in the class $-1$. Hence, the second and third constraints  force respectively the fraction of individuals with label $y_i=-1$ (respectively, $y_i=1$) counted as correctly classified to be at least $\lambda_{-1}^*$ (respectively, at least $\lambda_1^*$). Finally, the fourth constraint forces to select those features $k$ with $z_k=1.$
Note that, if very demanding classification rates are imposed, problem $(P1)$ may be infeasible. The solver will return this message, advising thus the user to lower the threshold values $\lambda_1$, $\lambda_{-1}$.

Solving $(P1)$ identifies the features to be used in the classification. However, an SVM classifier has not been built yet, since the margin has not been maximized. {The next section shall address such problem by using the SVM either with the linear kernel or with an arbitrary one.}

We should stress that the feature selection is based on the linear kernel, yielding the tractable linear integer optimization problem $(P1)$.
Extension of our FS approach to nonlinear kernels are formally straightforward, but the resulting nonconvex mixed integer nonlinear problems are not tractable, even for low dimensions.
For this reason, we perform the FS by assuming a linear kernel, and then, once the features are selected, the classifier is built using an arbitrary kernel, as detailed in Section~\ref{linvsarb}.

Of course more flexibility is gained if, in a preprocessing step, data $x$ are embedded in a higher dimensional space through a nonlinear mapping $\phi$, and thus the original $x$ is replaced by $\phi(x)$ in $(P1)$.

\subsection{Cost-sensitive sparse SVMs: linear vs arbitrary kernels}\label{linvsarb}
{Here we explain how the sparse SVM is built. Let us first consider the case of the classifier with linear kernel. Hence, the sparse SVM that controls the classification rates is formulated as}
 $$
\left.
\begin{array}{lll}
  \min_{w, \beta, \xi} &  \sum\limits_{j=1}^N{w_j^2 z_j} + C\sum_{i \in I} \xi_i& \\
  s.t. & y_i(\sum_{j=1}^N w_jz_j x_{ij} + \beta) \geq 1 - \xi_i,& \forall i \in I \\
   & 0 \leq \xi_i \leq M_1(1-\zeta_i) & \forall i \in I\\
   & \zeta_i \in \{0,1\}& \forall i \in I\\
    &  \sum_{i \in I} \zeta_i (1-y_i )  \geq \lambda_{-1}^* {\sum_{i \in I} (1-y_i)}&\\
 & \sum_{i  \in I} \zeta_i (1+y_i) \geq \lambda_{1}^* \sum_{i \in I}(1+y_i ).&\\
\end{array}
\right.
\qquad (P2)
$$

Note that ($P2$) is defined similarly as a standard linear SVM optimization problem. The slight difference is that in ($P2$) only the variables selected by the FS approach described in Section~\ref{FSProc}. are considered. This means that the values of $z$ in ($P2$) are those obtained in problem ($P1$). Note too that the constraints concerning the performance measures are also added here.

Now, assume the SVM classifier has the form (\ref{eq:score2}), and  an arbitrary kernel function $K(x,x')=\phi(x)^\top\phi(x')$ is used instead of the linear one.  See e.g. \cite{carrizosa2013supervised,cristianini2000introduction, vapkinnature, vapnik1998statistical} for details. Although formally similar, the case of an arbitrary kernel $K$ implies that{, if an FS procedure as ($P1$) is desired,} nonlinear constraints {are involved} and thus the optimization problem is much harder {to solve}. For this reason, instead of coping with {such hard problem, we propose an alternative strategy: first, $(P1)$ is solved {(as before)}, and then the SVM classifier (with the selected kernel) is built, using only the features selected in the problem described in Section~\ref{FSProc}.}
%
%
%
In what follows we focus on the radial kernel, {even though} one can consider any arbitrary kernel $K$. First, we define the binary variables $z$ identifying the features which are selected for classifying. The choice of the features, identified with  the vector $z$, leads to the kernel $K_z,$ defined as
$$
K_z(x,x')= exp\left( -\gamma \left(\sum\limits_{k=1}^{N} {z_k}(x^{(k)} - x'^{(k)})^2\right) \right),
$$
{where $x^{(k)}$ denotes the $k$-th component of vector $x$.}\\

For $z$ (and thus $K_z$) fixed, the aim is to solve (\ref{eq:svm2}), but replacing the terms $\boldsymbol{w}^\top \boldsymbol{w}$ and $\boldsymbol{w}^\top \phi(x_i)$, respectively, by the expressions
(\ref{eq:recovery}) and (\ref{eq:recoverscore}){, apart from adding the constraints related to the performance measurements, as described in \cite{CSVM}}. Therefore, the cost-sensitive sparse SVM with an arbitrary kernel $K$ is defined (once $z$ is fixed) as
%
%
%
%
%

$$
\left.
\begin{array}{lll}
\min_{\alpha,\mathbf{\xi},\beta,\mathbf{\zeta}} & \sum_{i,j \in I} \alpha_i y_i \alpha_{j} y_{j} K_z(x_i,x_{j}) + C \sum_{i \in I} \xi_i & \\
s.t. & y_i(\sum_{j \in I} \alpha_j y_j K_z(x_{i},x_{j}) + \beta) \geq 1 - \xi_i,& \forall i \in {I} \\
     & 0 \leq \xi_i \leq M_1(1-\zeta_i) & \forall i \in {I}  \\
     & \sum_{i \in I} \alpha_i y_i = 0\\
     & 0 \leq \alpha_i \leq C/2 & \forall i \in {I}  \\
       & { \sum\limits_{i \in I} \zeta_i (1-y_i) }\geq \lambda_{-1}^*{\sum\limits_{i \in I}(1-y_i)}  &\\
     & { \sum\limits_{i \in I} \zeta_i (1+y_i) } \geq \lambda_{1}^*{\sum\limits_{i \in I} (1+y_i)}&\\
     & \zeta_i \in \{0,1\} & \forall i \in I
\end{array}
\right.
\qquad (P3)
$$

Let us discuss the formulation $(P3)$. The set of  features is fixed through $z$. The objective function, the first, third and fourth constraints are the usual ones in SVM. The second constraint together with the fifth, sixth and seventh constraints force some samples to be correctly classified, as in $(P1).$



\section{Experiment Description}
 \label{Experimentdescription}

{In this section, a description of the experiments to be carried out in Section~\ref{Results} of the cost-sensitive sparse SVM with linear kernel (problem ($P2$)) are compared to those under the radial kernel (problem ($P3$)), where, as described in the previous section, the variables $z$ in both ($P2$) and ($P3$) are the solutions of the FS problem formulated by ($P1$). Also, the solutions under the sparse methodology will be tested against the standard linear SVM. {Although it would be natural to compare the solutions of ($P3$) with the solutions of a standard radial SVM, this comparison is not straightforward since ($P1$) may become infeasible when the performance measures obtained with the radial SVM are higher than those under the linear SVM. For simplicity we assume all measurement costs equal to 1, and then our aim is to minimize the number of features used.}}

{Next, a description of how the experiments have been carried out is given.} 
{In order to solve problems $(P1)$, $(P2)$ and $(P3)$, the solver Gurobi, \cite{gurobi}, and its Python language interface, \cite{pthn}, are used.}{ In order to estimate the performance of these FS procedures, a 10-fold cross-validation (CV), \cite{kohavi1995study}, is used, and out of samples accuracies are reported. {However, for those datasets that have less than 100 instances, a Leave-One-Out procedure is carried out, in order to have a good size in the training sample}.
{ Also, depending on whether the linear or the radial kernel is considered, a parameter $C$ or a pair of parameters ($C,\gamma$) must be tuned. Hence, in either the first or in the second case, {$C \in \{2^{-5},2^{-4},\ldots,2^4,2^5\}, \gamma \in \{2^{-5},2^{-4},\ldots,2^4,2^5\}$} are considered. Problems in integer variables are hard to solve to optimality. However, excellent solutions are obtained in reasonable time. {A time limit of 300 seconds is set, giving the solver enough time for finding (sub)optimal solutions. Parameters {$M_1$, $M_2$ and $M_3$} are set as 100. Moreover, parameters tuning is done by another 10-fold CV {(respectively, another Leave-One-Out)}, and the best set of parameters selected is the one with highest accuracy in average {(or, in the case of unbalanced data, with the highest geometric mean between the TPR and the TNR)}.}}

For a better understanding, the whole procedure is summarized in Algorithm~\ref{algFS2}.

\begin{algorithm}
    \SetKwInOut{Input}{Input}
    \SetKwInOut{Output}{Output}
    \For{kf = 1,$\ldots$,folds}{
      Split data ($D$) into ``\textit{folds}'' subsets, $ D=\{D_1,\ldots, D_{folds}\}$ \\
      Set $Validation = D_{kf}$ and set $I = D - \{ D_{kf} \}$\\
      \For{each pair $(C,gamma)$}{
          \For{kf2 = 1,$\ldots$, folds2}{
          split $D' = D - \{ D_{kf} \}$ into ``\textit{folds2}'' subsets, $ D'=\{ D'_1,\ldots, D'_{folds2}\}$ \\
          Set $Validation' = D'_{kf2}$ and set $I'  = D' - \{ D_{kf2}\}$\\
          Run ($P1$) over $I$, and select the relevant features.\\
          Run ($P2$) or ($P3$) over $I$ with the corresponding modified kernel.\\\
          Validate over $Validation'$, getting the accuracy ($acc[kf2]$)\\
          }
          Calculate the average accuracies $(\sum_{kf2} acc[kf2])/folds2$  \\
          \If{$acc[kf2] \geq bestacc$}{
                Set $bestacc$ = $acc[kf2]$, $bestgamma$ = $gamma$ and $bestC$ = $C$\\
          }
      }
     Run ($P1$) over $I$, and select the relevant features.\\
     Run ($P2$) or ($P3$) with the corresponding modified kernel and the parameters $bestgamma$ and $bestC$, using $I$.\\
      Validate over $Validation$, getting the accuracy ($acc2[kf]$), and the correct classification probabilities ($TPR[kf]$, $TNR[kf]$) as well as the number of features selected $Z[kf] = \sum_{k=1}^N z[k]$.
    }
    Calculate and display the average performance measures: $(\sum_{kf} acc2[k2])/folds$, $(\sum_{kf} TPR[kf])/folds$, $(\sum_{kf} TNR[kf])/folds$ and $(\sum_{kf} Z[kf])/folds$

    \caption{Pseudocode for general kernel approach.}\label{algFS2}
\end{algorithm}

\vspace{0.5cm}
\section{Numerical Results}
\label{Results}
{Here, the experimental results are presented. We have chosen the datasets \texttt{wisconsin} (Breast Cancer Wisconsin (Diagnostic) Data Set), \texttt{votes}{ (Congressional Voting Records Data Set), \texttt{nursery} (Nursery Data Set), \texttt{Australian} (Statlog (Australian Credit Approval) Data Set), \texttt{careval} (Car Evaluation Data Set) {and \texttt{gastrointestinal} (Gastrointestinal Lesions in Regular Colonoscopy Data Set),} all well referenced and described with detail in \cite{Lichman2013}{, and \texttt{leukemia} (Leukemia data), described in \cite{Golub531}.} {First, a brief data description is given in Section~\ref{DataDesc}}. Then, results under the linear kernel approach will be presented and discussed in Section \ref{ResLinear}. Finally, the case of the radial kernel will be analyzed in Section~\ref{ResRadial}.}

{Note that the main idea of a FS approach is to reduce the number of features or, more generally, the overall associated costs, in such a way that the performance is not severely affected. As we can control the proportion of samples well classified, this is not a problematic issue. In fact, experiments are done so that new performance measurements will not be {$0.025$} points lower than the originals ones, i.e., those obtained under the standard version of the SVM with linear kernel. {Using the notation as in \cite{CSVM}, $TNR$ and $TPR$ are the true negative and true positive rates, and $TNR_0$ and $TPR_0$ are their {obtained values under the standard SVM with linear kernel on a validation sample}, $TNR \geq \lambda_{-1} = \min\{1,TNR_0-0.025\}$ and $TPR \geq \lambda_1 = \min\{1,TPR_0-0.025\}$ {are desired. For both linear and radial cases we have considered the two possible selection of the thresholds, defined by (\ref{eq:HoeffornotNot}) and (\ref{eq:HoeffornotHoeff}).
}}}

We stress that the purpose of this experimental section is to show how we can control TPR or TNR without a severe deterioration of overall classification rates, hopefully with a strong decrease in the number or cost of the features selected. This is the reason why we are comparing the performance of our approach with respect to the performance of the standard SVM. In a real application, the thresholds $\lambda_1$, $\lambda_{-1}$ are to be given by the user, either based or not on SVM classification rates.

\subsection{Data description}
\label{DataDesc}
The performance of these novel approaches is illustrated using six real-life datasets from the UCI Repository,  \cite{Lichman2013}{{, as well as the \texttt{leukemia} dataset, \cite{Golub531}}. {The positive label will be assigned to the majority class in 2-class datasets. In addition,} multiclass datasets are transformed into 2-class ones, by giving positive label to the largest class and negative labels to the remaining samples. Categorical variables are transformed into dummy variables, i.e,  if a categorical variable with $\nu$ levels is present, it will be replaced by $\nu - 1$ binary variables. {Also, if there exist missing values, they are replaced by the median.} A description of the datasets can be found in Table~\ref{tab:data}. Such table is split in 4 columns. The first shows the name of the dataset {(the actual names of the datasets are presented at the beginning of this section)}. The total number of samples of the dataset is given in the second column. {The number of variables considered, and the number (and percentage) of positive samples in the dataset, are given in the last two columns.}

\begin{table}[h!]

\centering 

\begin{tabular}{llll}

     \hline
     Name           & $|\Omega|$ & $V$ & {$|\Omega_+|$} (\%)  \\
     \hline
    \texttt{wisconsin} & 569   & 30  & 357 (62.7 \%)\\
    \texttt{votes}   &  435  & 32 & 267 (61.4 \%) \\
    \texttt{nursery}  & 12960 & 19 & 4320 (33.3 \%)\\
    \texttt{Australian} & 690  & 34 & 383 (55.5 \%)\\
    \texttt{careval} & 1728   & 15 & 1210 (70.023 \%)\\
    \texttt{leukemia} &  72  & 7128 & 47 (65.278 \%)\\
    \texttt{gastrointestinal} &  76  & 698 & 55 (72.368 \%)\\
    \hline

 \end{tabular}

\caption{Details concerning the implementation of the CSVM for the considered datasets.}

\label{tab:data}

\end{table}

\subsection{Results under the cost-sensitive sparse SVM with linear kernel}\label{ResLinear}
{Two types of results will be shown, corresponding to the choices (\ref{eq:HoeffornotNot}) and (\ref{eq:HoeffornotHoeff}) of the thresholds. As a summary, it will turn out that (\ref{eq:HoeffornotNot}) yields sparser classifiers, while (\ref{eq:HoeffornotHoeff}), which is a more conservative choice, usually yields less sparse classifiers but with better accuracies.}

The choice of threshold parameters in (\ref{eq:HoeffornotNot}) leads to results summarized in Table~\ref{tab:LKA}. The first column of Table~\ref{tab:LKA} gives the name of the dataset used {(the abbreviation we have chosen for the dataset)}. Then, the second and third columns show, respectively, the performance measures for the standard SVM (using the linear kernel) and the proposed cost-sensitive sparse methodology. Such columns are split into two subcolumns: the first one shows the average values and the second one the standard deviations. The last column reports the feature reduction, by indicating the original and selected (average) number of variables.
\begin{table}[htbp]
  \centering
  \caption{Performance measures under the cost-sensitive sparse SVM with linear kernel and $\lambda_1^* = \lambda_1$, $\lambda_{-1}^* = \lambda_{-1}$.}
    \extrarowsep=_3pt^3pt
     \begin{tabu}to\linewidth{lllllllll}
    \tabucline[1pt gray]-
     Name & &  SVM &    & & FS &    &  & Feature reduction \\\cmidrule{3-4}\cmidrule{6-7}
        &  & Mean & Std& & Mean     & Std &  & \\
    \tabucline[1pt gray]-
    \texttt{wisconsin} & \texttt{Acc} & \texttt{0.975} & \texttt{0.021} & & \texttt{0.947} & \texttt{0.025} & & 30 $\rightarrow$ 2  (0 Std) \\[-8pt]
                   & \texttt{TPR} & \texttt{0.992} & \texttt{0.013} & & \texttt{0.973} & \texttt{0.031} \\[-8pt]
                   &\texttt{TNR} & \texttt{0.943} & \texttt{0.051} & & \texttt{0.905} & \texttt{0.063}\\
                   \hline
    \texttt{votes} & \texttt{Acc} & \texttt{0.954} & \texttt{0.033} & & \texttt{0.949} & \texttt{0.036} & & 32 $\rightarrow$ 2 (0 Std) \\[-8pt]
                   & \texttt{TPR} & \texttt{0.955} & \texttt{0.038} & & \texttt{0.928} & \texttt{0.059}\\[-8pt]
                   &\texttt{TNR} & \texttt{0.947} &  \texttt{0.059} & & \texttt{0.979} & \texttt{0.036}\\
                   \hline
    \texttt{nursery} & \texttt{Acc} & \texttt{1} & \texttt{0} & & \texttt{1} & \texttt{0} & &  19 $\rightarrow$  1 (0 Std) \\[-8pt]
                   & \texttt{TPR} & \texttt{1} & \texttt{0} & & \texttt{1} & \texttt{0}\\[-8pt]
                   &\texttt{TNR} & \texttt{1} &  \texttt{0} & & \texttt{1} & \texttt{0}\\
                   \hline
    \texttt{Australian} & \texttt{Acc} & \texttt{0.848} & \texttt{0.051} & & \texttt{0.855} & \texttt{0.057} & &  34 $\rightarrow$ 1 (0 Std) \\[-8pt]
                   & \texttt{TPR} & \texttt{0.798} & \texttt{0.083} & & \texttt{0.801} & \texttt{0.087}\\[-8pt]
                   &\texttt{TNR} & \texttt{0.912} &  \texttt{0.05} & & \texttt{0.926} & \texttt{0.041}\\
                   \hline
      \texttt{careval} & \texttt{Acc} & \texttt{0.956} & \texttt{0.017} & & \texttt{0.946} & \texttt{0.019} & &  15 $\rightarrow$ 9 (0 Std) \\[-8pt]
                   & \texttt{TPR} & \texttt{0.96} & \texttt{0.022} & & \texttt{0.963} & \texttt{0.017}\\[-8pt]
                   &\texttt{TNR} & \texttt{0.948} &  \texttt{0.024} & & \texttt{0.907} & \texttt{0.04}\\
                   \hline
      {\texttt{leukemia}} & {\texttt{Acc}} & {\texttt{0.972}} & {\texttt{0.164}} & & \texttt{0.875} & \texttt{0.331} & & 7128 $\rightarrow$ 3.139 (1.205 Std) \\[-8pt]
                   & {\texttt{TPR}} & {\texttt{0.979}} & {\texttt{0.196}} & & \texttt{0.896} & \texttt{0.305}\\[-8pt]
                   &{\texttt{TNR}} & {\texttt{0.96}} &  {\texttt{0.144}} & & \texttt{0.833} & \texttt{0.373}\\
                   \hline
      {\texttt{gastrointestinal}} & {\texttt{Acc}} &  {\texttt{0.895}} & {\texttt{0.307}} & & \texttt{0.829} & \texttt{0.379} & & 698  $\rightarrow$ 1 (0 Std) \\[-8pt]
                   & {\texttt{TPR}} &  {\texttt{0.929}} &  {\texttt{0.258}} & & \texttt{0.839} & \texttt{0.367}\\[-8pt]
                   &{\texttt{TNR}} &  {\texttt{0.8}} &   {\texttt{0.4}} & & \texttt{0.8} & \texttt{0.4}\\
                   \hline
    \tabucline[1pt gray]-
    \end{tabu}%
 \label{tab:LKA}%
\end{table}%
\begin{table}[htbp]
  \centering
  \caption{Performance measures under the cost-sensitive sparse SVM with linear kernel and $\lambda_1^* = \lambda_1 + \sqrt{-\log \alpha/(2|I_1|)}$,
  $\lambda_{-1}^* = \lambda_{-1} + \sqrt{-\log \alpha/(2|I_{-1}|)}$.}
    \extrarowsep=_3pt^3pt
     \begin{tabu}to\linewidth{lllllllll}
    \tabucline[1pt gray]-
     Name & &  SVM &    & & FS &    &  & Feature reduction \\\cmidrule{3-4}\cmidrule{6-7}
        &  & Mean & Std& & Mean     & Std &  & \\
    \tabucline[1pt gray]-
    \texttt{wisconsin} & \texttt{Acc} & \texttt{0.975} & \texttt{0.021} & & \texttt{0.965} & \texttt{0.023} & & 30 $\rightarrow$ 6.2  (0.919 Std) \\[-8pt]
                   & \texttt{TPR} & \texttt{0.992} & \texttt{0.013} & & \texttt{0.975} & \texttt{0.023} \\[-8pt]
                   &\texttt{TNR} & \texttt{0.943} & \texttt{0.051} & & \texttt{0.947} & \texttt{0.048}\\
                   \hline
    \texttt{votes} & \texttt{Acc} & \texttt{0.954} & \texttt{0.033} & & \texttt{0.954} & \texttt{0.033} & & 32 $\rightarrow$ 9.3 (1.16 Std) \\[-8pt]
                   & \texttt{TPR} & \texttt{0.955} & \texttt{0.038} & & \texttt{0.96} & \texttt{0.034}\\[-8pt]
                   &\texttt{TNR} & \texttt{0.947} &  \texttt{0.059} & & \texttt{0.945} & \texttt{0.052}\\
                   \hline
    \texttt{nursery} & \texttt{Acc} & \texttt{1} & \texttt{0} & & \texttt{1} & \texttt{0} & &  19 $\rightarrow$  1 (0 Std) \\[-8pt]
                   & \texttt{TPR} & \texttt{1} & \texttt{0} & & \texttt{1} & \texttt{0}\\[-8pt]
                   &\texttt{TNR} & \texttt{1} &  \texttt{0} & & \texttt{1} & \texttt{0}\\
                   \hline
    {\texttt{Australian}} & \texttt{Acc} & \texttt{0.848} & \texttt{0.051} & & \texttt{0.837} & \texttt{0.057} & &  34 $\rightarrow$ 5.5 (1.78 Std) \\[-8pt]
                   & \texttt{TPR} & \texttt{0.769} & \texttt{0.083} & & \texttt{0.772} & \texttt{0.074}\\[-8pt]
                   &\texttt{TNR} & \texttt{0.912} &  \texttt{0.05} & & \texttt{0.924} & \texttt{0.053}\\
                   \hline
      \texttt{careval} & \texttt{Acc} & \texttt{0.956} & \texttt{0.017} & & \texttt{0.954} & \texttt{0.018} & &  15 $\rightarrow$ 11 (0 Std) \\[-8pt]
                   & \texttt{TPR} & \texttt{0.96} & \texttt{0.022} & & \texttt{0.962} & \texttt{0.018}\\[-8pt]
                   &\texttt{TNR} & \texttt{0.948} &  \texttt{0.024} & & \texttt{0.935} & \texttt{0.039}\\
                   \hline
            {\texttt{leukemia}} & {\texttt{Acc}} & {\texttt{0.972}} & {\texttt{0.164}} & & \texttt{0.944} & \texttt{0.229} & & 7128 $\rightarrow$ 2 (0 Std) \\[-8pt]
                   & {\texttt{TPR}} & {\texttt{0.979}} & {\texttt{0.196}} & & \texttt{0.957} & \texttt{0.202}\\[-8pt]
                   &{\texttt{TNR}} & {\texttt{0.96}} &  {\texttt{0.144}} & & \texttt{0.92} & \texttt{0.272}\\
                   \hline
      {\texttt{gastrointestinal}} & {\texttt{Acc}} &  {\texttt{0.895}} & {\texttt{0.307}} & & \texttt{0.842} & \texttt{0.365} & & 698  $\rightarrow$ 3.105 (0.552 Std) \\[-8pt]
                   & {\texttt{TPR}} &  {\texttt{0.929}} &  {\texttt{0.258}} & & \texttt{0.927} & \texttt{0.26}\\[-8pt]
                   &{\texttt{TNR}} &  {\texttt{0.8}} &   {\texttt{0.4}} & & \texttt{0.619} & \texttt{0.486}\\
                   \hline
    \tabucline[1pt gray]-
    \end{tabu}%
 \label{tab:LKAH}%
\end{table}%
{From the table, it can be concluded that the approach with a linear kernel works well in general. In the case of \texttt{wisconsin}, the TPR has desirable values, since it only differentiates -0.019 points from the original. However, in the case of the accuracy and TNR, the loss is bigger than 0.025 points. This is due mainly to two reasons: first, the constraints are imposed on the training sample, while the performance is calculated using a test sample. Second, since the thresholds are considered as $\lambda_1^* = \lambda_1$, $\lambda_{-1}^*=\lambda_{-1}$, this implies we are not much restrictive as if $\lambda_1^* > \lambda_1$ ($\lambda_{-1}^*> \lambda_{-1}$) were required. Nevertheless, the new TNR value is only 0.038 points smaller than the original, and the reduction of features is significant since only two variables out of 30 are used. Also, in \texttt{votes} the features are significantly reduced and the most affected performance measure is the TPR, which decreases 0.027 points, making the accuracy smaller. However, the value on the TNR is increased. As happened with \texttt{wisconsin}, the loss is due mainly to the two facts previously mentioned. For \texttt{nursery}, an amazing reduction to only one feature is achieved, in addition getting a perfect classification. This is explained as follows. As commented in Section~\ref{DataDesc}, multiclass datasets are transformed into 2-class ones, and this is the case, obtaining the classes ``\texttt{not\_recom}'' and ``\texttt{others}'', which are the positive and negative classes, respectively. In addition, one of the (categorical) features in the data (which is the one selected by our procedure) completely determines the class. In \texttt{Australian}, the total number of variables is also reduced to only one, having similar performance measures values as in the standard SVM. In fact, we obtain here even better results than under the original linear SVM. If the variable selected with the algorithm is studied, one can observe that it is a binary variable $X$, where the contingency table together with the class variable is given in {Table~\ref{tab:conting}.} Hence this variable is by itself a good predictor, as the FS procedure pointed out.
 In the case of \texttt{careval}, we got the smallest reduction in the number of variables selected, maintaining the performance measures values above the imposed thresholds.
 {On the other hand, in the case of \texttt{leukemia}, the number of variables is significantly reduced. However, since the number of instances is small, the performance measurements are affected by this reduction of features.
 Also, for \texttt{gastrointestinal}, the results are similar to what happened for \texttt{leukemia}, but the TNR has not been affected at all.}
\begin{tabular}{ccc}

\end{tabular}

\begin{table}[h!]

\centering 

\begin{tabular}{l|ll}

   & X = 0 & X = 1 \\
   \hline
  Class $+$& 306 & 77 \\
  Class $-$& 23 & 284

 \end{tabular}

\caption{Contingency table of the feature selected in \texttt{Australian}.}

\label{tab:conting}

\end{table}
}

{Consider next the results shown by Table~\ref{tab:LKAH}, for the case where we are restrictive regarding the performance values, that is, when $\lambda_1^* = \lambda_1 + \sqrt{-\log \alpha/(2|I_1|)}$ and
  $\lambda_{-1}^* = \lambda_{-1} + \sqrt{-\log \alpha/(2|I_{-1}|)}$. From the table, it can be seen how this approach tends to work better concerning the performance measures, but achieves less sparse solutions. For example, if we focus on \texttt{wisconsin}, the TNR, the TPR and the accuracy as well, obtain the desired performance requirements. However, only a reduction of variables of one fifth is obtained. In the case of \texttt{votes}, an analogous result is obtained for the performance measures and only a reduction in one third of the variables is achieved. The same pattern as before is observed for \texttt{nursery}. For \texttt{Australian}, we obtain even an improvement in all the three performance measures considered, reducing the number of features to one fifth. {In addition}, we get again in \texttt{careval} the smallest reduction in the number of variables selected, maintaining the performance measures values above the thresholds imposed as before, but using a larger number of features. {On the contrary, and surprisingly, we have obtained for \texttt{leukemia} even a bigger reduction in the number of features and better results using Hoeffding inequality. However, \texttt{gastrointestinal} dataset goes with the flow and the number of features is increased when using the mentioned inequality. Nevertheless, the TPR has not been affected now whereas the TNR has significantly decreased.}}

\subsection{Results under the cost-sensitive sparse SVM with radial kernel}\label{ResRadial}
{The analogous results to those in Section~\ref{ResLinear} are presented here, for the case of the radial kernel. {However, only \texttt{wisconsin}, \texttt{votes} and \texttt{Australian} datasets are used here.} As shown by Tables~\ref{tab:acc} and \ref{tab:acc2} and similarly as occurred in Section~\ref{ResLinear}, the use of the threshold values obtained by the Hoeffding inequality (as in (\ref{eq:HoeffornotHoeff})) tends to yield a lower level of sparsity, but also, a higher predictive power in general {(particularly, when achieving the desired bounds)}. Concerning the performance measures, it can be deduced from Tables~\ref{tab:acc} and \ref{tab:acc2} that {this approach works well in general, especially when using Hoeffding}. Finally, it should be noted how the reduction in the number of features is quite notable for some datasets, {as before}. }

\begin{table}[htbp]
  \centering
  \caption{Performance measures under the cost-sensitive sparse SVM with radial kernel and $\lambda_1^* = \lambda_1$, $\lambda_{-1}^* = \lambda_{-1}$.}
     \extrarowsep=_3pt^3pt
     \begin{tabu}to\linewidth{lllllllll}
    \tabucline[1pt gray]-
     Name & &  SVM &    & & FS &    &  & Feature reduction \\\cmidrule{3-4}\cmidrule{6-7}
        &  & Mean & Std& & Mean     & Std &  & \\
    \tabucline[1pt gray]-
    {\texttt{wisconsin}} & \texttt{Acc} & \texttt{0.975} & \texttt{0.021} & & \texttt{0.956} & \texttt{0.012} & & 30 $\rightarrow$ 2  (0 Std) \\
                   & \texttt{TPR} & \texttt{0.992} & \texttt{0.013} & & \texttt{0.988} & \texttt{0.016} \\
                   &\texttt{TNR} & \texttt{0.943} & \texttt{0.051} & & \texttt{0.893} & \texttt{0.051}\\
                   \hline
    {\texttt{votes}} & \texttt{Acc} & \texttt{0.954} & \texttt{0.033} & & \texttt{0.947} & \texttt{0.034} & & 32 $\rightarrow$ 2 (0 Std) \\
                   & \texttt{TPR} & \texttt{0.955} & \texttt{0.038} & & \texttt{0.928} & \texttt{0.059}\\
                   &\texttt{TNR} & \texttt{0.947} &  \texttt{0.059} & & \texttt{0.974} & \texttt{0.036}\\
                   \hline
    \texttt{nursery} & \texttt{Acc} & \texttt{1} & \texttt{0} & & \texttt{1} & \texttt{0} & &  19 $\rightarrow$  1 (0 Std) \\
                   & \texttt{TPR} & \texttt{1} & \texttt{0} & & \texttt{1} & \texttt0{}\\
                   &\texttt{TNR} & \texttt{1} &  \texttt{0} & & \texttt{1} & \texttt{0}\\
                   \hline
    \tabucline[1pt gray]-
    \end{tabu}%
 \label{tab:acc}%
\end{table}%

\begin{table}[htbp]
  \centering
  \caption{Performance measures under the cost-sensitive sparse SVM with radial kernel and $\lambda_1^* = \lambda_1 + \sqrt{-\log \alpha/(2|I_1|)}$,
  $\lambda_{-1}^* = \lambda_{-1} + \sqrt{-\log \alpha/(2|I_{-1}|)}$.}
     \extrarowsep=_3pt^3pt
     \begin{tabu}to\linewidth{lllllllll}
    \tabucline[1pt gray]-
     Name & &  SVM &    & & FS &    &  & Feature reduction \\\cmidrule{3-4}\cmidrule{6-7}
        &  & Mean & Std& & Mean     & Std &  & \\
    \tabucline[1pt gray]-
     {\texttt{wisconsin}} & \texttt{Acc} & \texttt{0.975} & \texttt{0.021} & & \texttt{0.947} & \texttt{0.03} & & 30 $\rightarrow$ 6.2 (0.919 Std) \\
                   & \texttt{TPR} & \texttt{0.992} & \texttt{0.013} & & \texttt{0.967} & \texttt{0.039} \\
                   &\texttt{TNR} & \texttt{0.943} & \texttt{0.051} & & \texttt{0.907} & \texttt{0.02}\\
                   \hline
    {\texttt{votes}} & \texttt{Acc} & \texttt{0.954} & \texttt{0.033} & & \texttt{0.949} & \texttt{0.03} & & 32 $\rightarrow$ 9.3 (1.16 Std) \\
                   & \texttt{TPR} & \texttt{0.955} & \texttt{0.038} & & \texttt{0.959} & \texttt{0.034}\\
                   &\texttt{TNR} & \texttt{0.947} &  \texttt{0.059} & & \texttt{0.939} & \texttt{0.043}\\
                   \hline
    \texttt{nursery} & \texttt{Acc} & \texttt{1} & \texttt{0} & & \texttt{1} & \texttt{0} & &  19 $\rightarrow$  1 (0 Std) \\
                   & \texttt{TPR} & \texttt{1} & \texttt{0} & & \texttt{1} & \texttt{0}\\
                   &\texttt{TNR} & \texttt{1} &  \texttt{0} & & \texttt{1} & \texttt{0}\\
                   \hline
    \tabucline[1pt gray]-
    \end{tabu}%
 \label{tab:acc2}%
\end{table}%

\newpage

\subsection{{Comparison with other methodologies}}\label{Comparative}

{The cost-sensitive FS procedure presented here can be compared in a certain way with some other benchmark methodologies. However, the authors are not aware of FS methods for SVM controlling, as we do, TPR or TNR. Among the different FS techniques that can be applied to SVM we can find, for example, the following ones: \textbf{Filter methods} (they are based on measures like Pearson's Correlation, Linear Discriminant Analysis or Chi-Square), \textbf{Wrapper methods} (Forward Selection, Backward Elimination, Recursive Feature elimination, \ldots) and \textbf{Embedded methods} (such as the presented in the Introduction Section).}

{In order to make a comparison with another state-of-the-art method, we have selected the method in \cite{Chan2007DCR12734961273515},\cite{GHADDAR2018993}. The results can be seen in Tables~\ref{tab:LKAComp} and \ref{tab:LKAComp2}. In Table~\ref{tab:LKAComp} we can see the results for the standard SVM, the results of our FS approach when $\lambda_1^* = \lambda_1$ and $\lambda_{-1}^* = \lambda_{-1}$, and the results of the state-of-the-art method when the maximum number of features selected is the same as the obtained for our methodology. In Table~\ref{tab:LKAComp2}, the results for the standard SVM are reported together with the results of our FS approach when $\lambda_1^* = \lambda_1 + \sqrt{-\log \alpha/(2|I_1|)}$ and $\lambda_{-1}^* = \lambda_{-1} + \sqrt{-\log \alpha/(2|I_{-1}|)}$, as well as the results of the state-of-the-art method when the maximum number of features selected is the same as the obtained with our methodology.}

\begin{table}[htbp]
  \centering
  \caption{Performance measures under the cost-sensitive sparse SVM with linear kernel and $\lambda_1^* = \lambda_1$, $\lambda_{-1}^* = \lambda_{-1}$ and comparative with the method in \cite{Chan2007DCR12734961273515},\cite{GHADDAR2018993}.}
    \extrarowsep=_3pt^3pt
     \begin{tabu}to\linewidth{llllllllll}
    \tabucline[1pt gray]-
     Name & &  SVM &    & & FS &    &  & Compar.  &  \\\cmidrule{3-4}\cmidrule{6-7}\cmidrule{9-10}
        &  & Mean & Std& & Mean     & Std &  &  Mean     & Std \\
    \tabucline[1pt gray]-
    \texttt{wisconsin} & \texttt{Acc} & \texttt{0.975} & \texttt{0.021} & & \texttt{0.947} & \texttt{0.025} & & 0.954 & \texttt{0.021}\\[-8pt]
                   & \texttt{TPR} & \texttt{0.992} & \texttt{0.013} & & \texttt{0.973} & \texttt{0.031} & & \texttt{0.977} & \texttt{0.025}\\[-8pt]
                   &\texttt{TNR} & \texttt{0.943} & \texttt{0.051} & & \texttt{0.905} & \texttt{0.063} & & \texttt{0.911} & \texttt{0.056}\\
                   \hline
    \texttt{votes} & \texttt{Acc} & \texttt{0.954} & \texttt{0.033} & & \texttt{0.949} & \texttt{0.036} & & \texttt{0.956} & \texttt{0.026}\\[-8pt]
                   & \texttt{TPR} & \texttt{0.955} & \texttt{0.038} & & \texttt{0.928} & \texttt{0.059} & & \texttt{0.949} & \texttt{0.039}\\[-8pt]
                   &\texttt{TNR} & \texttt{0.947} &  \texttt{0.059} & & \texttt{0.979} & \texttt{0.036} & & \texttt{0.969} & \texttt{0.034}\\
                   \hline
    \texttt{nursery} & \texttt{Acc} & \texttt{1} & \texttt{0} & & \texttt{1} & \texttt{0} & & \texttt{1} & \texttt{0}  \\[-8pt]
                   & \texttt{TPR} & \texttt{1} & \texttt{0} & & \texttt{1} & \texttt{0}& & \texttt{1} & \texttt{0}\\[-8pt]
                   &\texttt{TNR} & \texttt{1} &  \texttt{0} & & \texttt{1} & \texttt{0} & & \texttt{1} & \texttt{0}\\
                   \hline
    \texttt{Australian} & \texttt{Acc} & \texttt{0.848} & \texttt{0.051} & & \texttt{0.855} & \texttt{0.057} & & \texttt{0.855} & \texttt{0.054}\\[-8pt]
                   & \texttt{TPR} & \texttt{0.798} & \texttt{0.083} & & \texttt{0.801} & \texttt{0.087} & & \texttt{0.801} & \texttt{0.082}\\[-8pt]
                   &\texttt{TNR} & \texttt{0.912} &  \texttt{0.05} & & \texttt{0.926} & \texttt{0.041} & & \texttt{0.925} & \texttt{0.039}\\
                   \hline
      \texttt{careval} & \texttt{Acc} & \texttt{0.956} & \texttt{0.017} & & \texttt{0.946} & \texttt{0.019} & &  \texttt{0.949} & \texttt{0.016}\\[-8pt]
                   & \texttt{TPR} & \texttt{0.96} & \texttt{0.022} & & \texttt{0.963} & \texttt{0.017} & & \texttt{0.967} & \texttt{0.012}\\[-8pt]
                   &\texttt{TNR} & \texttt{0.948} &  \texttt{0.024} & & \texttt{0.907} & \texttt{0.04} & & \texttt{0.91} & \texttt{0.043} \\
                   \hline
      \texttt{leukemia} & \texttt{Acc} & \texttt{0.972} & \texttt{0.164} & & \texttt{0.875} & \texttt{0.331} & &  \texttt{0.653} & \texttt{0.471}\\[-8pt]
                   & \texttt{TPR} & \texttt{0.979} & \texttt{0.196} & & \texttt{0.896} & \texttt{0.305} & & \texttt{0.66} & \texttt{0.474}\\[-8pt]
                   &\texttt{TNR} & \texttt{0.96} &  \texttt{0.144} & & \texttt{0.833} & \texttt{0.373} & & \texttt{0.68} & \texttt{0.466} \\
                   \hline
      \texttt{gastrointestinal} & \texttt{Acc} & \texttt{0.895} & \texttt{0.307} & & \texttt{0.829} & \texttt{0.379} & &  \texttt{0.857} & \texttt{0.35}\\[-8pt]
                   & \texttt{TPR} & \texttt{0.929} & \texttt{0.258} & & \texttt{0.839} & \texttt{0.367} & & \texttt{0.9} & \texttt{0.3}\\[-8pt]
                   &\texttt{TNR} & \texttt{0.8} &  \texttt{0.4} & & \texttt{0.8} & \texttt{0.4} & & \texttt{0.75} & \texttt{0.433} \\
                   \hline
    \tabucline[1pt gray]-
    \end{tabu}%
 \label{tab:LKAComp}%
\end{table}%

\begin{table}[htbp]
  \centering
  \caption{Performance measures under the cost-sensitive sparse SVM with linear kernel and $\lambda_1^* = \lambda_1 + \sqrt{-\log \alpha/(2|I_1|)}$,
  $\lambda_{-1}^* = \lambda_{-1} + \sqrt{-\log \alpha/(2|I_{-1}|)}$ and comparative with the method in \cite{Chan2007DCR12734961273515},\cite{GHADDAR2018993}.}
    \extrarowsep=_3pt^3pt
     \begin{tabu}to\linewidth{llllllllll}
    \tabucline[1pt gray]-
     Name & &  SVM &    & & FS &    &  & Compar.  & \\\cmidrule{3-4}\cmidrule{6-7}\cmidrule{9-10}
        &  & Mean & Std& & Mean     & Std &  &  Mean     & Std \\
    \tabucline[1pt gray]-
    \texttt{wisconsin} & \texttt{Acc} & \texttt{0.975} & \texttt{0.021} & & \texttt{0.965} & \texttt{0.023} & & \texttt{0.967} & \texttt{0.018} \\[-8pt]
                   & \texttt{TPR} & \texttt{0.992} & \texttt{0.013} & & \texttt{0.975} & \texttt{0.023} & & \texttt{0.989} & \texttt{0.017}\\[-8pt]
                   &\texttt{TNR} & \texttt{0.943} & \texttt{0.051} & & \texttt{0.947} & \texttt{0.048} & & \texttt{0.926} & \texttt{0.033}\\
                   \hline
    \texttt{votes} & \texttt{Acc} & \texttt{0.954} & \texttt{0.033} & & \texttt{0.954} & \texttt{0.033} & & \texttt{0.954} & \texttt{0.033} \\[-8pt]
                   & \texttt{TPR} & \texttt{0.955} & \texttt{0.038} & & \texttt{0.96} & \texttt{0.034} & & \texttt{0.948} & \texttt{0.036}\\[-8pt]
                   &\texttt{TNR} & \texttt{0.947} &  \texttt{0.059} & & \texttt{0.945} & \texttt{0.052} & & \texttt{0.961} & \texttt{0.035}\\
                   \hline
    \texttt{nursery} & \texttt{Acc} & \texttt{1} & \texttt{0} & & \texttt{1} & \texttt{0} & & \texttt{1} & \texttt{0}  \\[-8pt]
                   & \texttt{TPR} & \texttt{1} & \texttt{0} & & \texttt{1} & \texttt{0}& & \texttt{1} & \texttt{0}\\[-8pt]
                   &\texttt{TNR} & \texttt{1} &  \texttt{0} & & \texttt{1} & \texttt{0} & & \texttt{1} & \texttt{0}\\
                   \hline
    \texttt{Australian} & \texttt{Acc} & \texttt{0.848} & \texttt{0.051} & & \texttt{0.837} & \texttt{0.057} & & \texttt{0.851} & \texttt{0.053}\\[-8pt]
                   & \texttt{TPR} & \texttt{0.798} & \texttt{0.083} & & \texttt{0.772} & \texttt{0.074} & & \texttt{0.798} & \texttt{0.081}\\[-8pt]
                   &\texttt{TNR} & \texttt{0.912} &  \texttt{0.05} & & \texttt{0.924} & \texttt{0.053} & & \texttt{0.919} & \texttt{0.046}\\
                   \hline
      \texttt{careval} & \texttt{Acc} & \texttt{0.956} & \texttt{0.017} & & \texttt{0.954} & \texttt{0.018} & & \texttt{0.954}  & \texttt{0.017} \\[-8pt]
                   & \texttt{TPR} & \texttt{0.96} & \texttt{0.022} & & \texttt{0.962} & \texttt{0.018} & & \texttt{0.97} & \texttt{0.016}\\[-8pt]
                   &\texttt{TNR} & \texttt{0.948} &  \texttt{0.024} & & \texttt{0.935} & \texttt{0.039} & & \texttt{0.917} & \texttt{0.027} \\
                   \hline
      \texttt{leukemia} & \texttt{Acc} & \texttt{0.972} & \texttt{0.164} & & \texttt{0.944} & \texttt{0.229} & &  \texttt{0.932} & \texttt{0.252}\\[-8pt]
                   & \texttt{TPR} & \texttt{0.979} & \texttt{0.196} & & \texttt{0.957} & \texttt{0.202} & & \texttt{0.938} & \texttt{0.242}\\[-8pt]
                   &\texttt{TNR} & \texttt{0.96} &  \texttt{0.144} & & \texttt{0.92} & \texttt{0.272} & & \texttt{0.917} & \texttt{0.276} \\
                   \hline
      \texttt{gastrointestinal} & \texttt{Acc} & \texttt{0.895} & \texttt{0.307} & & \texttt{0.842} & \texttt{0.365} & &  \texttt{0.714} & \texttt{0.452}\\[-8pt]
                   & \texttt{TPR} & \texttt{0.929} & \texttt{0.258} & & \texttt{0.927} & \texttt{0.26} & & \texttt{0.75} & \texttt{0.433}\\[-8pt]
                   &\texttt{TNR} & \texttt{0.8} &  \texttt{0.4} & & \texttt{0.619} & \texttt{0.486} & & \texttt{0.625} & \texttt{0.484} \\
                   \hline
    \tabucline[1pt gray]-
    \end{tabu}%
 \label{tab:LKAComp2}%
\end{table}%

{We can observe that, {except for \texttt{gastrointestinal} dataset (when using Hoeffding inequality), where we obtain better results than the comparative}, similar results are obtained for our method and the method in \cite{Chan2007DCR12734961273515},\cite{GHADDAR2018993} in terms of accuracy, while our methodology is cost-sensitive and we can control the performance measurements. As an illustration, in Table~\ref{tab:allFeat}  we have collected all the results for dataset \texttt{australian} when applying the method in \cite{Chan2007DCR12734961273515}, \cite{GHADDAR2018993} and varying the number of features from $1$ (minimum) to $34$ (maximum). There, we can see how the maximum TPR obtained is 0.8007, so with our methodology, and maybe at the expense of increasing misclassification rates in the another class, we can improve the accuracy rates in the target class. The results obtained when either TPR or TNR are varied are summarized in Table~\ref{tab:improveAllFeat}}.
}

\begin{table}[htbp]
  \centering
  \caption{Performance measures using the method in \cite{Chan2007DCR12734961273515},\cite{GHADDAR2018993}, varying the maximum number of features from 1 (minimum) to 34 (maximum) in \texttt{Australian} dataset.}
    \extrarowsep=_3pt^3pt
     \begin{tabu}to\linewidth{lllllllll}
     \tabucline[1pt gray]-
     Australian\\
    \cmidrule{1-2}
       \# Feat. & Acc &  TPR  & TNR & &   \# Feat. & Acc &  TPR  & TNR\\
      \cmidrule{1-4}\cmidrule{6-9}
       1 & 0.8551 & 0.8007 & 0.9248 & & 18 & 0.8464 & 0.7954 & 0.9121 \\[-4pt]
       2 & 0.8551 & 0.8007 & 0.9248 & & 19 & 0.8464 & 0.7954 & 0.9121 \\[-4pt]
       3 & 0.8551 & 0.8007 & 0.9248 & & 20 & 0.8464 & 0.7954 & 0.9121 \\[-4pt]
       4 & 0.8551 & 0.8007 & 0.9248 & & 21 & 0.8464 & 0.7954 & 0.9121 \\[-4pt]
       5 & 0.8507 & 0.798 & 0.9186 & & 22 & 0.8578 & 0.7953 & 0.9154\\[-4pt]
       6 & 0.8507 & 0.798 & 0.9186 & & 23 & 0.8464 & 0.7954 & 0.9121\\[-4pt]
       7 & 0.8507 & 0.798 & 0.9186 & & 24 & 0.8464 & 0.7954 & 0.9121\\[-4pt]
       8 & 0.8507 & 0.798 & 0.9186 & & 25 & 0.8464 & 0.7954 & 0.9121\\[-4pt]
       9 & 0.8507 & 0.798 & 0.9186 & & 26 & 0.8449 & 0.7929 & 0.9121\\[-4pt]
       10 & 0.8507 & 0.798 & 0.9186 & & 27 & 0.8478 & 0.7981 & 0.9121\\[-4pt]
       11 & 0.8507 & 0.798 & 0.9186 & & 28 & 0.8478 & 0.7954 & 0.9153\\[-4pt]
       12 & 0.8478 & 0.798 & 0.9121 & & 29 & 0.8464 & 0.7927 & 0.9153 \\[-4pt]
       13 & 0.8478 & 0.798 & 0.9121 & & 30 & 0.8493 & 0.798 & 0.9153 \\[-4pt]
       14 & 0.8478 & 0.798 & 0.9121 & & 31 & 0.8478 & 0.7954 & 0.9153 \\[-4pt]
       15 & 0.8478 & 0.798 & 0.9121 & & 32 & 0.8478 & 0.7981 & 0.9121\\[-4pt]
       16 & 0.8464 & 0.7954 & 0.9121 & & 33 & 0.8478 & 0.7981 & 0.9121\\[-4pt]
       17 & 0.8478 & 0.798 & 0.9121 & & 34 & 0.8478 & 0.7981 & 0.9121\\[0pt]
    \tabucline[1pt gray]-
    \end{tabu}%
 \label{tab:allFeat}%
\end{table}%

\begin{table}[htbp]
  \centering
  \caption{Performance measures using the method in \cite{Chan2007DCR12734961273515},\cite{GHADDAR2018993}, varying the maximum number of features from 1 (minimum) to 34 (maximum) in \texttt{Australian} dataset.}
    \extrarowsep=_3pt^3pt
     \begin{tabu}to\linewidth{cccccc}
     \tabucline[1pt gray]-
     Australian\\
    \cmidrule{1-2}
        $\lambda_{1}^*$ & $\lambda_{-1}^*$ & Acc &  TPR  & TNR & Aver. \# Feat. Selected\\
    \tabucline[1pt gray]-
       0.85 & 0.5 & 0.738 & 0.94 & 0.484 & 1 \\
       0.85 & 0.55 & 0.738 & 0.94 & 0.484 & 1 \\
       0.85 & 0.575 & 0.812 & 0.854 & 0.754 & 1.7\\
       0.85 & 0.6 & 0.855 & 0.801 & 0.92 & 2\\
       0.9 & 0.5 & 0.757 & 0.896 & 0.582 & 1.333\\
    \tabucline[1pt gray]-
    \end{tabu}%
 \label{tab:improveAllFeat}%
\end{table}%

{From these experimental results we can conclude that, indeed, our method is able not only to reduce the number of features but it also controls the performance measures. If we observe the cases where ($\lambda_{1}^*$,$\lambda_{-1}^*$) is (0.85,0.5) or (0.85,0.55), we see how we have strongly increased the value in the TPR, although the TNR has decreased a lot. A similar behavior is observed for the pair (0.9,0.5). However, when using (0.85,0.575) a different trade-off is found, whereas for (0.85,0.6) we recover the original results.}

\section{Concluding remarks}
\label{se:concluding}
In this paper we have proposed a Feature Selection procedure for Support Vector Machines that yields a novel, sparse, SVM. Contrary to existing Feature Selection approaches, we take explicitly into account that misclassification costs may be rather different in the two groups, and thus, instead of seeking the classifier maximizing the margin, we seek the most sparse classifier that attains {certain} true positive and true negative rates on the dataset.
{For both SVM with linear and radial kernel, the problem is written in a straightforward manner, solving first a mixed integer linear problem and then their standard SVM formulations, considering only the features obtained in the first problem as well as the performance constraints. The reported numerical results show that the novel approaches lead to comparable or better performance rates, in addition to an important reduction in the number of variables.}

Several extensions of the approach presented in this paper are possible. In our opinion, they deserve further study.
First, several classification and regression procedures based on optimization problems, such as Support Vector Regression, logistic regression or distance-weighted discrimination, are amenable to address, as done here, an integrated FS and classification or regression. The optimization problems obtained in this way have a structure which should be exploited to make the approach competitive and including cost-sensitivity in the FS procedure. Second, even within SVM, it should be observed that SVM is a tool for binary classification. For multiclass datasets, SVM classification is performed by solving a series of SVM problems, see \cite{cristianini2000introduction,wang2007}. When some classes are hard to identify,  the basic multiclass strategies may yield discouraging results. Performing simultaneously feature selection and class fusion, as in \cite{guo2010}, is an interesting nontrivial extension of our approach. To do this, problems (P1), (P2) and (P3) need to be conveniently modified.


\section*{Acknowledgements}
This research is financed by Fundaci\'on BBVA, projects FQM329 and P11-FQM-7603 (Junta de Andaluc\'{\i}a, Andaluc\'{\i}a) and  MTM2015-65915-R (Ministerio de Econom\'{\i}a y Competitividad, Spain). The last three are cofunded with EU ERD Funds. The authors are thankful for such support.

\section*{Appendix}
{
In this section we describe step by step how formulation~(\ref{eq:csvm2}) is built from equation~(\ref{eq:csvm}). Hence, let us suppose first that we have the model
\begin{equation*}
\begin{array}{lll}
  \min_{\boldsymbol{w}, \beta, \xi} & \boldsymbol{w}^\top \boldsymbol{w}+ C\sum_{i \in I} \xi_i& \\
  s.t. & y_i(\boldsymbol{w}^\top x_i + \beta) \geq 1 - \xi_i,& i \in I \\
   & 0 \leq \xi_i \leq M_1(1-\zeta_i)& i \in I\\
   & \mu(\zeta)_\ell \geq \lambda_\ell & \ell \in L\\
   & \zeta_i \in \{0,1\} & i \in I.
\end{array}
\end{equation*}
This one can be rewritten as
\[
\begin{array}{lllllll}
\min_{{\zeta}}   & & &  {\min_{{\boldsymbol{w}},\beta,{\xi}} }& {{\boldsymbol{w}}^\top {\boldsymbol{w}}  + C \sum\limits_{i \in  I } \xi_i }\\
\mbox{s.t.} & \zeta_i \in \{0,1\} & i \in I  &  {\mbox{s.t.}}  &y_i\left(\boldsymbol{w}^\top x_i + \beta\right) \geq 1 - \xi_i,& i \in I \\
   & \mu(\zeta)_\ell \geq \lambda_\ell & \ell \in L &   & 0 \leq \xi_i \leq M_1(1-\zeta_i)& i \in I\\
\end{array}
\]
If we assume that the binary variables $\zeta$ fixed, the Karush--K{u}hn--Tucker (KKT) conditions for the inner problem are
\[
\begin{array}{llll}
{w} & = & \sum\limits_{i \in I} \alpha_i y_i {x}_i \\
0 & = & \sum\limits_{i \in I} \alpha_i y_i \\
0 & \leq & \alpha_i \leq C/2  & i\in I. \\
\end{array}
\]
Substituting these expressions into the last optimization problem, the partial dual of such problem can be calculated, obtaining
\[
\begin{array}{lllll}
\min\limits_{{\zeta}}   & & &  {\min\limits_{{\alpha}, \beta, {\xi}} } & {\left(  \sum\limits_{i \in I} \alpha_i y_i {x}_i\right)^\top
\left(  \sum\limits_{i \in I} \alpha_i y_i {x}_i \right)  + C \sum\limits_{i \in  I } \xi_i }\\
\mbox{s.t.} & z_j \in \{0,1\} & j \in J  & {\mbox{s.t.}}  & {y_i \left( \left(\sum\limits_{i \in I} \alpha_i y_i {x}_i \right)^\top {x}_i  + \beta \right)
 \geq 1 - \xi_i} \quad {i \in I}\\
 & \mu(\zeta)_\ell \geq \lambda_\ell & \ell \in L & & {0 \leq \xi_i \leq M_1(1-\zeta_i)} \quad {i \in I }\\
& & & & {\sum\limits_{i \in I} \alpha_i y_i  = 0}\\
& &  &  & { 0 \leq \alpha_i \leq C/2} \quad {i \in I }\\
\end{array}
\]
As a last step, the kernel trick is used and the final formulation~(\ref{eq:csvm2}) is obtained.
}

\section*{References}

\bibliography{FS_SVM_COR}

\begin{thebibliography}{}

\bibitem[Aytug, 2015]{aytug2015feature}
Aytug, H. (2015).
\newblock Feature selection for support vector machines using {G}eneralized
  {B}enders {D}ecomposition.
\newblock {\em European Journal of Operational Research}, 244(1):210--218.

\bibitem[Bartlett et~al., 2006]{bartlett2006convexity}
Bartlett, P.~L., Jordan, M.~I., and McAuliffe, J.~D. (2006).
\newblock Convexity, classification, and risk bounds.
\newblock {\em Journal of the American Statistical Association},
  101(473):138--156.

\bibitem[Ben-Tal et~al., 2011]{ben2011chance}
Ben-Tal, A., Bhadra, S., Bhattacharyya, C., and Saketha~Nath, J. (2011).
\newblock Chance constrained uncertain classification via robust optimization.
\newblock {\em Mathematical Programming}, 127(1):145--173.

\bibitem[Ben\'itez-Pe\~na et~al., ]{CSVM}
Ben\'itez-Pe\~na, S., Blanquero, R., Carrizosa, E., and Ram\'irez-Cobo, P.
\newblock {On Support Vector Machines under a multiple-cost scenario}.
\newblock Working Paper.

\bibitem[Bertolazzi et~al., 2016]{bertolazzi2016integer}
Bertolazzi, P., Felici, G., Festa, P., Fiscon, G., and Weitschek, E. (2016).
\newblock Integer programming models for feature selection: {N}ew extensions
  and a randomized solution algorithm.
\newblock {\em European Journal of Operational Research}, 250(2):389--399.

\bibitem[Bertsimas et~al., 2016a]{bertsimas2016best}
Bertsimas, D., King, A., Mazumder, R., et~al. (2016a).
\newblock Best subset selection via a modern optimization lens.
\newblock {\em The Annals of Statistics}, 44(2):813--852.

\bibitem[Bertsimas et~al., 2014]{bertsimas2014least}
Bertsimas, D., Mazumder, R., et~al. (2014).
\newblock Least quantile regression via modern optimization.
\newblock {\em The Annals of Statistics}, 42(6):2494--2525.

\bibitem[Bertsimas et~al., 2016b]{bertsimas2016tae}
Bertsimas, D., O'Hair, A.~K., and Pulleyblank, W.~R. (2016b).
\newblock {\em The Analytics Edge}.
\newblock Dynamic Ideas, Massachusetts.

\bibitem[Bo{\c{t}} and Lorenz, 2011]{boct2011optimization}
Bo{\c{t}}, R.~I. and Lorenz, N. (2011).
\newblock Optimization problems in statistical learning: Duality and optimality
  conditions.
\newblock {\em European Journal of Operational Research}, 213(2):395--404.

\bibitem[Bradley et~al., 1999]{bradley1999mathematical}
Bradley, P.~S., Fayyad, U.~M., and Mangasarian, O.~L. (1999).
\newblock Mathematical {P}rogramming for {D}ata {M}ining: {F}ormulations and
  {C}hallenges.
\newblock {\em INFORMS Journal on Computing}, 11(3):217--238.

\bibitem[Bradley et~al., 1998]{bradley1998feature}
Bradley, P.~S., Mangasarian, O.~L., and Street, W.~N. (1998).
\newblock Feature {S}election via {M}athematical {P}rogramming.
\newblock {\em INFORMS Journal on Computing}, 10(2):209--217.

\bibitem[Carrizosa et~al., 2008]{carrizosa2008multi}
Carrizosa, E., Mart\'{\i}n-Barrag\'an, B., and Romero-Morales, D. (2008).
\newblock Multi-group support vector machines with measurement costs: A
  biobjective approach.
\newblock {\em Discrete Applied Mathematics}, 156(6):950--966.

\bibitem[Carrizosa et~al., 2011]{carrizosa2011detecting}
Carrizosa, E., Mart{\'\i}n-Barrag{\'a}n, B., and {Romero-Morales}, D. (2011).
\newblock Detecting relevant variables and interactions in supervised
  classification.
\newblock {\em European Journal of Operational Research}, 213(1):260--269.

\bibitem[Carrizosa et~al., 2016]{carrizosa2016strongly}
Carrizosa, E., Nogales-G{\'o}mez, A., and {Romero-Morales}, D. (2016).
\newblock Strongly agree or strongly disagree?: Rating features in support
  vector machines.
\newblock {\em Information Sciences}, 329:256--273.

\bibitem[Carrizosa et~al., 2017a]{carrizosa2017clustering}
Carrizosa, E., Nogales-G{\'o}mez, A., and {Romero-Morales}, D. (2017a).
\newblock Clustering categories in support vector machines.
\newblock {\em Omega}, 66:28--37.

\bibitem[Carrizosa et~al., 2017b]{carrizosa2016sparsity}
Carrizosa, E., Olivares-Nadal, A.~V., and Ram{\'\i}rez-Cobo, P. (2017b).
\newblock A sparsity-controlled vector autoregressive model.
\newblock {\em Biostatistics}, page kxw042.

\bibitem[Carrizosa and {Romero-Morales}, 2013]{carrizosa2013supervised}
Carrizosa, E. and {Romero-Morales}, D. (2013).
\newblock Supervised classification and mathematical optimization.
\newblock {\em Computers \& Operations Research}, 40(1):150--165.

\bibitem[Chan et~al., 2007]{Chan2007DCR12734961273515}
Chan, A.~B., Vasconcelos, N., and Lanckriet, G. R.~G. (2007).
\newblock Direct convex relaxations of sparse svm.
\newblock In {\em Proceedings of the 24th International Conference on Machine
  Learning}, ICML '07, pages 145--153, New York, NY, USA. ACM.

\bibitem[Corne et~al., 2012]{corne2012synergies}
Corne, D., Dhaenens, C., and Jourdan, L. (2012).
\newblock Synergies between operations research and data mining: The emerging
  use of multi-objective approaches.
\newblock {\em European Journal of Operational Research}, 221(3):469--479.

\bibitem[Cristianini and Shawe-Taylor, 2000]{cristianini2000introduction}
Cristianini, N. and Shawe-Taylor, J. (2000).
\newblock {\em An {I}ntroduction to {S}upport {V}ector {M}achines and {O}ther
  {K}ernel-based {L}earning {M}ethods}.
\newblock Cambridge {U}niversity {P}ress.

\bibitem[Fung and Mangasarian, 2004]{fung2004feature}
Fung, G.~M. and Mangasarian, O.~L. (2004).
\newblock A {F}eature {S}election {N}ewton {M}ethod for {S}upport {V}ector
  {M}achine {C}lassification.
\newblock {\em Computational Optimization and Applications}, 28(2):185--202.

\bibitem[Ghaddar and Naoum-Sawaya, 2018]{GHADDAR2018993}
Ghaddar, B. and Naoum-Sawaya, J. (2018).
\newblock High dimensional data classification and feature selection using
  support vector machines.
\newblock {\em European Journal of Operational Research}, 265(3):993 -- 1004.

\bibitem[Golub et~al., 1999]{Golub531}
Golub, T.~R., Slonim, D.~K., Tamayo, P., Huard, C., Gaasenbeek, M., Mesirov,
  J.~P., Coller, H., Loh, M.~L., Downing, J.~R., Caligiuri, M.~A., Bloomfield,
  C.~D., and Lander, E.~S. (1999).
\newblock Molecular classification of cancer: Class discovery and class
  prediction by gene expression monitoring.
\newblock {\em Science}, 286(5439):531--537.

\bibitem[Guo, 2010]{guo2010}
Guo, J. (2010).
\newblock Simultaneous variable selection and class fusion for high-dimensional
  linear discriminant analysis.
\newblock {\em Biostatistics}, 11(4):599.

\bibitem[{Gurobi Optimization, Inc.}, 2016]{gurobi}
{Gurobi Optimization, Inc.} (2016).
\newblock Gurobi optimizer reference manual.

\bibitem[Guyon and Elisseeff, 2003]{guyon2003introduction}
Guyon, I. and Elisseeff, A. (2003).
\newblock An {I}ntroduction to {V}ariable and {F}eature {S}election.
\newblock {\em Journal of {M}achine {L}earning {R}esearch}, 3(Mar):1157--1182.

\bibitem[Kohavi, 1995]{kohavi1995study}
Kohavi, R. (1995).
\newblock {A Study of Cross-Validation and Bootstrap for Accuracy Estimation
  and Model Selection}.
\newblock In {\em {IJCAI}}, volume~14, pages 1137--1143. Stanford, CA.

\bibitem[Le~Thi et~al., 2015]{le2015feature}
Le~Thi, H.~A., Le, H.~M., and Dinh, T.~P. (2015).
\newblock Feature selection in machine learning: an exact penalty approach
  using a {D}ifference of {C}onvex function {A}lgorithm.
\newblock {\em Machine Learning}, 101(1):163--186.

\bibitem[Lichman, 2013]{Lichman2013}
Lichman, M. (2013).
\newblock {UCI} {M}achine {L}earning {R}epository.

\bibitem[Maldonado et~al., 2017]{MALDONADO2017656}
Maldonado, S., P\'erez, J., and Bravo, C. (2017).
\newblock Cost-based feature selection for support vector machines: An
  application in credit scoring.
\newblock {\em European Journal of Operational Research}, 261(2):656 -- 665.

\bibitem[Maldonado and Weber, 2009a]{maldonado2009wrapper}
Maldonado, S. and Weber, R. (2009a).
\newblock A wrapper method for feature selection using {S}upport {V}ector
  {M}achines.
\newblock {\em Information Sciences}, 179(13):2208--2217.

\bibitem[Maldonado and Weber, 2009b]{MALDONADO20092208}
Maldonado, S. and Weber, R. (2009b).
\newblock A wrapper method for feature selection using support vector machines.
\newblock {\em Information Sciences}, 179(13):2208 -- 2217.
\newblock Special Section on High Order Fuzzy Sets.

\bibitem[Maldonado et~al., 2011a]{maldonado2011simultaneous}
Maldonado, S., Weber, R., and Basak, J. (2011a).
\newblock Simultaneous feature selection and classification using
  kernel-penalized support vector machines.
\newblock {\em Information Sciences}, 181(1):115--128.

\bibitem[Maldonado et~al., 2011b]{MALDONADO2011115}
Maldonado, S., Weber, R., and Basak, J. (2011b).
\newblock Simultaneous feature selection and classification using
  kernel-penalized support vector machines.
\newblock {\em Information Sciences}, 181(1):115 -- 128.

\bibitem[Marron et~al., 2007]{marron2007distance}
Marron, J.~S., Todd, M.~J., and Ahn, J. (2007).
\newblock Distance-weighted discrimination.
\newblock {\em Journal of the American Statistical Association},
  102(480):1267--1271.

\bibitem[Meisel and Mattfeld, 2010]{meisel2010synergies}
Meisel, S. and Mattfeld, D. (2010).
\newblock Synergies of {O}perations {R}esearch and {D}ata {M}ining.
\newblock {\em European Journal of Operational Research}, 206(1):1--10.

\bibitem[Panagopoulos et~al., 2016]{panagopoulos2016constrained}
Panagopoulos, O.~P., Pappu, V., Xanthopoulos, P., and Pardalos, P.~M. (2016).
\newblock Constrained subspace classifier for high dimensional datasets.
\newblock {\em Omega}, 59:40--46.

\bibitem[Plastria and Carrizosa, 2012]{plastria2012minmax}
Plastria, F. and Carrizosa, E. (2012).
\newblock Minmax-distance approximation and separation problems: geometrical
  properties.
\newblock {\em Mathematical {P}rogramming}, 132(1):153--177.

\bibitem[Provost and Fawcett, 2013]{provost2013data}
Provost, F. and Fawcett, T. (2013).
\newblock {\em Data {S}cience for {B}usiness: {W}hat {Y}ou {N}eed to {K}now
  about {D}ata {M}ining and {D}ata-{A}nalytic {T}hinking}.
\newblock O'Reilly Media, Inc., 1st edition.

\bibitem[{Python Core Team}, 2015]{pthn}
{Python Core Team} (2015).
\newblock {Python: A dynamic, open source programming language}.
\newblock Python Software Foundation.

\bibitem[Richt{\'a}rik and Tak{\'a}{\v{c}}, 2016]{richtarik2016parallel}
Richt{\'a}rik, P. and Tak{\'a}{\v{c}}, M. (2016).
\newblock Parallel coordinate descent methods for big data optimization.
\newblock {\em Mathematical {P}rogramming}, 156(1):433--484.

\bibitem[S\'anchez et~al., 2016]{kxw018}
S\'anchez, B.~N., Wu, M., Song, P. X.~K., and Wang, W. (2016).
\newblock Study design in high-dimensional classification analysis.
\newblock {\em Biostatistics}, 17(4):722.

\bibitem[Shen et~al., 2003]{shen2003psi}
Shen, X., Tseng, G.~C., Zhang, X., and Wong, W.~H. (2003).
\newblock On $\psi$-learning.
\newblock {\em Journal of the American Statistical Association},
  98(463):724--734.

\bibitem[Vapnik, 1995]{vapkinnature}
Vapnik, V. (1995).
\newblock {\em The {N}ature of {S}tatistical {L}earning {T}heory}.
\newblock Springer-Verlag New York, Inc., New York, NY, USA.

\bibitem[Vapnik, 1998]{vapnik1998statistical}
Vapnik, V. (1998).
\newblock {\em Statistical learning theory}, volume~1.
\newblock Wiley New York.

\bibitem[Wang and Shen, 2007]{wang2007}
Wang, L. and Shen, X. (2007).
\newblock {On L1-Norm Multiclass Support Vector Machines}.
\newblock {\em Journal of the American Statistical Association},
  102(478):583--594.

\bibitem[Weston et~al., 2001]{weston2000feature}
Weston, J., Mukherjee, S., Chapelle, O., Pontil, M., Poggio, T., and Vapnik, V.
  (2001).
\newblock Feature {S}election for {SVM}s.
\newblock In Leen, T.~K., Dietterich, T.~G., and Tresp, V., editors, {\em
  Advances in Neural Information Processing Systems 13}, pages 668--674. MIT
  Press.

\end{thebibliography}

\end{document}